\documentclass[11pt]{article}
\ifdefined\ARRREVIEW
  \usepackage[review]{acl}
\else
  \usepackage[preprint]{acl}
\fi
\usepackage[T1]{fontenc}
\usepackage[utf8]{inputenc}
\usepackage{booktabs}
\usepackage{graphicx}
\usepackage{amsmath}
\usepackage{enumitem}
\usepackage{array}

\title{Known Intents, New Combinations:\\Clause-Factorized Decoding for Compositional Multi-Intent Detection}
\author{Abhilash Nandy\\Microsoft Research India\\\texttt{abnandy@microsoft.com}}
\date{}

\begin{document}
\maketitle

\begin{abstract}
Multi-intent detection papers usually ask whether a model can recover multiple intents from one utterance. We ask a harder and, for deployment, more useful question: can it recover \emph{new combinations of familiar intents}? Existing benchmarks only weakly test this, because train and test often share the same broad co-occurrence patterns. We introduce \textsc{CoMIX-Shift}, a controlled benchmark built to stress compositional generalization in multi-intent detection through held-out intent pairs, discourse-pattern shift, longer and noisier wrappers, held-out clause templates, and zero-shot triples. We also present \textsc{ClauseCompose}, a lightweight decoder trained only on singleton intents, and compare it to whole-utterance baselines including a fine-tuned tiny BERT model. Across three random seeds, \textsc{ClauseCompose} reaches 95.7 exact match on unseen intent pairs, 93.9 on discourse-shifted pairs, 62.5 on longer/noisier pairs, 49.8 on held-out templates, and 91.1 on unseen triples. WholeMultiLabel reaches 81.4, 55.7, 18.8, 15.5, and 0.0; the BERT baseline reaches 91.5, 77.6, 48.9, 11.0, and 0.0. We also add a 240-example manually authored SNIPS-style compositional set with five held-out pairs; there, \textsc{ClauseCompose} reaches 97.5 exact match on unseen pairs and 86.7 under connector shift, compared with 41.3 and 10.4 for WholeMultiLabel. The results suggest that multi-intent detection needs more compositional evaluation, and that simple factorization goes surprisingly far once evaluation asks for it.
\end{abstract}

\section{Introduction}
Multi-intent detection has improved steadily, but its evaluation target is still too easy. Many current benchmarks ask models to recover multiple intents from a single utterance, yet they do so under splits where the same kinds of intent combinations appear in both training and test. A model can look strong in that setup while mostly memorizing familiar co-occurrence patterns.

That is not how deployed assistants fail. They fail when users compose known capabilities in unfamiliar ways: ``cancel my meeting with Sam, and after that play Phoebe Bridgers'' should not become difficult simply because that exact pair never appeared in training. The problem is not label scarcity. The problem is composition.

This paper studies that missing axis directly: compositional generalization for multi-intent detection. In semantic parsing, compositional holdouts are now a standard way to distinguish true structural generalization from strong in-distribution pattern matching \citep{cui2022compositional, zheng2023grammar}. Multi-intent detection has a similar structure, but a weaker evaluation culture. \citet{jiang2023spm} already note that existing models struggle when test utterances contain more intents than training exposed. We take that observation one step further and make composition itself the evaluation variable.

Our starting point is straightforward. A whole-utterance model is encouraged to learn surface statistics over complete intent sets. That is useful when train and test share the same combinations, but brittle when they do not. A clause-factorized model, in contrast, can learn atomic intents once and reuse them compositionally at inference time. To test that hypothesis, we build \textsc{CoMIX-Shift}, a controlled benchmark with ten intents, explicit pair holdouts, discourse-pattern shifts, longer noisy wrappers, held-out clause templates, and zero-shot triples. We compare four systems:
\begin{enumerate}[leftmargin=*, itemsep=2pt]
    \item \textbf{AtomicSet}, which treats each seen intent set as a single class;
    \item \textbf{WholeMultiLabel}, a whole-utterance multi-label model with a learned count head;
    \item \textbf{BERTTinyMultiLabel}, a fine-tuned tiny BERT encoder with the same multi-label and count objective;
    \item \textbf{ClauseCompose}, our singleton-trained clause-factorized decoder.
\end{enumerate}

The result is sharp. On unseen triples with discourse shift, both whole-utterance baselines collapse to 0.0 exact match, while \textsc{ClauseCompose} retains 91.1. On unseen pairs, the BERT baseline is competitive at 91.5, but \textsc{ClauseCompose} still reaches 95.7. As soon as discourse, length, or lexical realization shift, the gap widens: on longer/noisier pairs, \textsc{ClauseCompose} leads the strongest whole-utterance baseline by 13.6 points; on held-out clause templates, it leads by 34.3 points.

We do not argue that synthetic evaluation should replace natural benchmarks. We argue for something narrower: if multi-intent detection is supposed to support productive recombination of known intents, then evaluation should measure that capability explicitly. Once it does, simple compositional structure becomes much more valuable than it looks under standard splits.

Our contributions are:
\begin{enumerate}[leftmargin=*, itemsep=2pt]
    \item We introduce \textsc{CoMIX-Shift}, a controlled benchmark for compositional multi-intent detection with held-out pairs, discourse shift, longer/noisier wrappers, held-out clause templates, and zero-shot triples.
    \item We propose \textsc{ClauseCompose}, a lightweight clause-factorized decoder trained only on singleton intents.
    \item We add two stronger checks that the first draft lacked: a fine-tuned tiny BERT baseline and a 240-example manually authored SNIPS-style compositional set with explicit held-out pairs.
    \item We show that compositional evaluation changes the model ranking: whole-utterance baselines remain strong on easy splits, while the factorized decoder dominates once composition, discourse, or lexical form shift.
\end{enumerate}

\section{Related Work}
Joint intent detection and slot filling has evolved from coupled recurrent models \citep{wang2018bimodel, e2019bidirectional} to graph-based multi-intent systems \citep{qin2020agif, qin2021glgin}, label-aware few-shot formulations \citep{hou2021fewshot}, and more recent co-attention models \citep{pham2023misca}. These models improve sentence-level accuracy on standard MID benchmarks, but they are still mostly optimized as whole-utterance predictors.

The closest prior work to our perspective is Split-Parsing Method (SPM) \citep{jiang2023spm}, which argues that decomposing a multi-intent utterance into single-intent sub-sentences improves generalization and slot assignment. Our work is complementary, but narrower and more controlled. We study intent-set prediction only, isolate compositional generalization as the main variable, and evaluate a simpler clause-factorized decoder in a benchmark where pair overlap, discourse shift, and arity shift can be varied independently.

Our benchmark design is motivated by compositional generalization work in semantic parsing. \citet{cui2022compositional} and \citet{zheng2023grammar} show that even strong seq2seq models remain brittle once train and test differ in the combinations of known components they contain. The same lesson applies here: if compositional generalization is part of the intended capability, it should appear in the split design rather than only in the motivation section.

\section{\textsc{CoMIX-Shift}: A Controlled Benchmark}
\subsection{Data Generation}
\textsc{CoMIX-Shift} is generated from ten atomic intents:
\texttt{check\_weather},
\texttt{weather\_alert},
\texttt{book\_restaurant},
\texttt{restaurant\_hours},
\texttt{restaurant\_address},
\texttt{play\_music},
\texttt{pause\_music},
\texttt{send\_message},
\texttt{schedule\_meeting}, and
\texttt{cancel\_meeting}.

Each intent is associated with five natural-language templates and a small slot lexicon for cities, people, cuisines, times, artists, and dates. Singleton utterances are sampled by filling templates with random slot values and light politeness noise (for example, optional ``please'' or ``for me''). Multi-intent utterances are created by composing singleton clauses with surface patterns such as ``$a$ and $b$'', ``please $a$, then $b$'', or discourse-shifted forms such as ``before anything else, $a$; once that is done, $b$''.

\subsection{Split Design}
The benchmark contains 45 possible intent pairs. We randomly assign 27 to the \emph{seen-pair} pool and hold out 18 as \emph{unseen pairs}. Training uses singleton examples and seen pairs only; no triples appear in training. To probe lexical transfer, we also reserve one singleton template per intent and never expose it during training. We evaluate on six test settings:
\begin{enumerate}[leftmargin=*, itemsep=2pt]
    \item \textbf{Seen pairs}: familiar pairs with familiar discourse patterns.
    \item \textbf{Unseen pairs}: held-out intent pairs with familiar discourse patterns.
    \item \textbf{Pair shift}: held-out intent pairs with discourse connectors not used in training.
    \item \textbf{Long shift}: held-out intent pairs wrapped in longer and noisier request scaffolds.
    \item \textbf{Template holdout}: held-out intent pairs whose component clauses use singleton templates not seen in training.
    \item \textbf{Triple shift}: zero-shot triples with shifted discourse patterns.
\end{enumerate}

Table~\ref{tab:data} summarizes the scale used in our experiments.

\begin{table}[t]
\centering
\small
\setlength{\tabcolsep}{5pt}
\resizebox{\columnwidth}{!}{
\begin{tabular}{lrr}
\toprule
Split & Examples & Notes \\
\midrule
Train single & 2,800 & 280 per intent \\
Train pair seen & 6,480 & 27 seen pairs \\
Test pair seen & 1,080 & familiar pairs \\
Test pair unseen & 1,080 & held-out pairs \\
Test pair shift & 1,080 & held-out pairs + new discourse \\
Test pair long shift & 1,080 & held-out pairs + longer/noisier wrappers \\
Test pair template & 1,080 & held-out pairs + held-out singleton templates \\
Test triple shift & 600 & unseen triples + new discourse \\
\bottomrule
\end{tabular}
}
\caption{Benchmark scale.}
\label{tab:data}
\end{table}

\section{Method}
\subsection{Baselines}
\paragraph{AtomicSet.}
This model encodes the utterance once and predicts one label from the inventory of intent sets seen during training. It is a deliberately brittle baseline: perfect memorization on seen pairs is possible, but unseen pairs and triples are out of class by construction.

\paragraph{WholeMultiLabel.}
This model also encodes the utterance once, but predicts a sigmoid score for each intent plus a three-way count head for predicting whether the utterance expresses one, two, or three intents. At test time, we take the top-$k$ intents under the predicted count.

\subsection{\textsc{ClauseCompose}}
\textsc{ClauseCompose} is intentionally lightweight. We first train a single-intent encoder on singleton utterances only. At inference time, we segment the test utterance with a small discourse-marker grammar, encode each segment independently, and score intents with the single-intent classifier. A greedy decoder then builds the final set by selecting one high-scoring intent per segment while discouraging duplicates.

The model hard-codes a structural prior: multi-intent utterances are treated as compositions of single-intent clauses. The prior is crude, but it buys two useful properties. First, the model can handle intent arities unseen in training because decoding scales with the number of segments rather than the inventory of training combinations. Second, it is less sensitive to whether the whole utterance resembles a familiar pair template.

\section{Experimental Setup}
AtomicSet, WholeMultiLabel, and ClauseCompose use the same mean-pooled token encoder with 96-dimensional embeddings and two projection layers. We train them with Adam, a learning rate of $2 \times 10^{-3}$, batch size 64, and 12 epochs. BERTTinyMultiLabel uses a locally cached \texttt{prajjwal1/bert-tiny} encoder with the same multi-label and count objective, fine-tuned for three epochs with AdamW at $2 \times 10^{-5}$. Results are averaged over three random seeds. The non-BERT pipeline runs on CPU in a few minutes, which makes the benchmark suitable for rapid iteration and ablation work.

We report exact-match accuracy over the predicted intent set, micro-F1, macro-F1, and intent-count accuracy. Exact match is the most informative metric here because it penalizes both missing and spurious intents, so the main paper tables focus on EM while the full metrics remain in the released artifacts.

We also report two additional checks. \textbf{ClauseComposeOracle} reuses the same singleton classifier as \textsc{ClauseCompose}, but replaces heuristic segmentation with gold clause boundaries at test time. This tells us whether the remaining errors come from clause classification or from the rule-based segmenter. Separately, we build a small \textbf{manual SNIPS-style compositional set}: 240 manually authored two-clause utterances over six intents, with five held-out intent pairs and a connector-shifted variant.

\section{Results}
Table~\ref{tab:main} reports the main results. All models solve the seen-pair split, which is unsurprising: once the pair inventory is familiar, even a brittle atomic classifier can memorize it. The interesting behavior begins as soon as the composition changes.

\begin{table*}[t]
\centering
\small
\setlength{\tabcolsep}{4pt}
\resizebox{\textwidth}{!}{
\begin{tabular}{lcccccc}
\toprule
Model & Seen & Unseen & Pair shift & Long shift & Template holdout & Triple shift \\
\midrule
AtomicSet & 100.0 $\pm$ 0.0 & 0.0 $\pm$ 0.0 & 0.0 $\pm$ 0.0 & 0.0 $\pm$ 0.0 & 0.0 $\pm$ 0.0 & 0.0 $\pm$ 0.0 \\
WholeMultiLabel & 100.0 $\pm$ 0.0 & 81.4 $\pm$ 7.6 & 55.7 $\pm$ 11.0 & 18.8 $\pm$ 15.3 & 15.5 $\pm$ 8.6 & 0.0 $\pm$ 0.0 \\
BERTTinyMultiLabel & 100.0 $\pm$ 0.0 & 91.5 $\pm$ 10.4 & 77.6 $\pm$ 20.0 & 48.9 $\pm$ 23.4 & 11.0 $\pm$ 9.2 & 0.0 $\pm$ 0.0 \\
\textbf{ClauseCompose} & 95.2 $\pm$ 1.9 & \textbf{95.7 $\pm$ 2.9} & \textbf{93.9 $\pm$ 3.3} & \textbf{62.5 $\pm$ 9.2} & \textbf{49.8 $\pm$ 17.2} & \textbf{91.1 $\pm$ 2.6} \\
\bottomrule
\end{tabular}
}
\caption{Main synthetic results over three seeds. All numbers are exact-match set accuracy (EM).}
\label{tab:main}
\end{table*}

\subsection{What Changes Under Composition Shift?}
Two patterns stand out. First, \textbf{whole-utterance memorization is sufficient for seen pairs}. Second, \textbf{the same strategy becomes unstable as soon as composition shifts}. WholeMultiLabel stays respectable on unseen pairs, but its exact match drops by 25.7 points under connector shift, another 36.9 points under longer/noisier wrappers, another 3.4 under held-out templates, and reaches 0.0 on unseen triples because its count head never learns to emit three intents. The tiny BERT baseline is stronger on the easier compositional splits, but it still falls to 48.9 on long/noisy pairs, 11.0 on held-out templates, and 0.0 on unseen triples. \textsc{ClauseCompose}, by contrast, gives up a few points on the easy split and stays strong where the combinatorics change.

That trade-off is the paper's central point. Under in-distribution evaluation, a whole-utterance model looks competitive and sometimes ideal. Under compositional evaluation, the ranking flips. The question is not ``which architecture is best in general?'' The question is ``what capability does the benchmark actually reward?''

\subsection{Manual Compositional Check}
Table~\ref{tab:manual} adds a small manually authored SNIPS-style check. This set is still modest, but it fixes the main problem with the earlier scratch SNIPS section: the evaluation now contains true cross-intent pair holdouts rather than two clauses from the same coarse label. The pattern from the synthetic benchmark survives. On held-out pairs, \textsc{ClauseCompose} reaches 97.5 EM while WholeMultiLabel reaches 41.3 and the BERT baseline reaches 16.7. Under connector shift, the factorized model still reaches 86.7, while both whole-utterance baselines sit near 10.

\begin{table}[t]
\centering
\small
\setlength{\tabcolsep}{5pt}
\resizebox{\columnwidth}{!}{
\begin{tabular}{lcc}
\toprule
Model & Manual unseen pairs & Manual pair shift \\
\midrule
WholeMultiLabel & 41.3 $\pm$ 3.1 & 10.4 $\pm$ 1.2 \\
BERTTinyMultiLabel & 16.7 $\pm$ 7.7 & 10.8 $\pm$ 6.8 \\
\textbf{ClauseCompose} & \textbf{97.5 $\pm$ 0.0} & \textbf{86.7 $\pm$ 7.4} \\
\bottomrule
\end{tabular}
}
\caption{Manual SNIPS-style compositional check over six intents and five held-out pairs. All numbers are EM.}
\label{tab:manual}
\end{table}

\subsection{Robustness and Segmentation}
The new long/noisy split addresses a second concern: perhaps the gains only hold when the utterance is neatly factorized. That split is materially harder for every non-oracle model. WholeMultiLabel drops to 18.8 EM and the BERT baseline to 48.9, while \textsc{ClauseCompose} retains 62.5. Oracle segmentation reaches 100.0 on both pair-shift and long-shift splits, but only 49.4 on held-out templates. So connector and wrapper noise mostly hurt the heuristic segmenter, whereas held-out templates expose a different bottleneck: lexical transfer in the singleton classifier.

\subsection{Qualitative Behavior}
Consider the utterance:
\begin{quote}
\small
``before anything else, start some lofi music by the weeknd; once that is done, cancel my meeting with priya''
\end{quote}
The pair itself is unseen and the discourse pattern is absent from training. A whole-utterance model must decide whether this new surface form still looks like some familiar intent set. \textsc{ClauseCompose} does something simpler: it isolates the two clauses and applies the singleton classifier twice. This is not deep reasoning, but it is the right bias for the test condition we care about.

\section{Discussion}
The benchmark is synthetic, and that matters. We do not claim that \textsc{CoMIX-Shift} captures all of real assistant traffic. What it captures cleanly is one failure mode that real systems still face: recombining known intents under new discourse realizations. The long/noisy split and the held-out-template split make that point less toy-like: even after adding more conversational scaffolding or removing familiar clause templates, explicit factorization still helps, though the gap narrows and variance grows. In that setting, a small amount of explicit structure goes a long way, but it is not sufficient for robust lexical generalization.

There is also a broader lesson for MID evaluation. Existing work often introduces more expressive interaction layers and reports gains on standard train/test splits \citep{qin2021glgin, pham2023misca}. Those gains may be real. But without compositional holdouts, it is hard to tell whether the model learned intent structure or merely learned better statistics over familiar combinations. Our tiny BERT baseline sharpens that point: it is strong on seen and mildly shifted pairs, yet it still collapses on triples and held-out templates. The manual SNIPS-style set points the same way, even though it is much smaller.

\section{Limitations}
This study has four clear limitations. First, the main benchmark is still templatic and synthetic. It isolates composition cleanly, but natural user language is messier. The new held-out-template and long/noisy splits reduce some of that gap, but they do not replace a real compositional benchmark. Second, our method uses a hand-written discourse-marker inventory, which is a useful bias here but not a complete segmentation strategy for real dialogue. Third, we evaluate set prediction only and leave slot filling aside, even though slot-intent alignment is central to deployed SLU systems.

A fourth limitation is that our non-synthetic evidence is still small and manually authored. The strongest next step would combine real mixed-intent data with explicit pair, template, and arity holdouts, so that the benchmark pressure better matches deployment conditions.

\section{Conclusion}
We introduced \textsc{CoMIX-Shift}, a controlled benchmark for compositional multi-intent detection, and \textsc{ClauseCompose}, a lightweight clause-factorized decoder trained only on singleton intents. The main takeaway is simple: when evaluation asks models to recover unseen combinations of familiar intents, explicit factorization becomes far more valuable than rich whole-utterance pattern matching. We hope this gives multi-intent detection a cleaner and more honest target for future evaluation.

\bibliography{references}

\end{document}